\documentclass[10pt,twocolumn,letterpaper]{article}

\usepackage{cvpr}
\usepackage{times}
\usepackage{epsfig}
\usepackage{graphicx}
\usepackage{amsmath}
\usepackage{amssymb}
\usepackage{booktabs}
\usepackage{tabulary}
\usepackage{xcolor}
\usepackage[caption=false]{subfig}
\usepackage{multirow}

\usepackage[pagebackref=true,breaklinks=true,letterpaper=true,colorlinks,bookmarks=false]{hyperref}

\cvprfinalcopy 

\def\cvprPaperID{****} 

\ifcvprfinal\fi

\newcommand{\secvspace}{\vspace{-0.2em}}
\newcommand{\subsecvspace}{\vspace{-0.2em}}

\newcommand{\app}{\raise.3ex\hbox{$\scriptstyle\sim$}}
\newcommand{\ftrans}{\mathcal{T}}
\newcommand{\thetatrans}{\theta}
\newcommand{\wseg}{w_\textrm{seg}}
\newcommand{\wcseg}{w_{\textrm{seg}}^\textit{c}}
\newcommand{\wdet}{w_\textrm{det}}
\newcommand{\wcdet}{w_{\textrm{det}}^\textit{c}}
\newcommand{\wccls}{w_{\textrm{cls}}^\textit{c}}
\newcommand{\wcbox}{w_{\textrm{box}}^\textit{c}}

\newcommand{\methodname}{Mask$^{\mathit{X}}$ R-CNN\xspace}

\newcommand{\myparagraph}[1]{\noindent\textbf{#1}}

\newlength\savewidth\newcommand\shline{\noalign{\global\savewidth\arrayrulewidth
  \global\arrayrulewidth 1pt}\hline\noalign{\global\arrayrulewidth\savewidth}}
\newcommand{\tablestyle}[2]{\setlength{\tabcolsep}{#1}\renewcommand{\arraystretch}{#2}\centering\footnotesize}
\newcolumntype{x}[1]{>{\centering\arraybackslash}p{#1pt}}
\makeatletter\renewcommand\myparagraph{\@startsection{paragraph}{4}{\z@}
  {.5em \@plus1ex \@minus.2ex}{-.5em}{\normalfont\normalsize\bfseries}}\makeatother
\definecolor{demphcolor}{RGB}{124,124,124}
\newcommand{\demph}[1]{\textcolor{demphcolor}{#1}}

\begin{document}

\title{Learning to Segment Every Thing}

\author{
Ronghang Hu$^{1,2,*}$ \quad Piotr Doll\'ar$^2$ \quad Kaiming He$^2$ \quad Trevor Darrell$^1$ \quad Ross Girshick$^2$ \vspace{.5em} \\
$^1$BAIR, UC Berkeley  \qquad $^2$Facebook AI Research (FAIR)
}

\maketitle
\renewcommand*{\thefootnote}{\fnsymbol{footnote}}
\setcounter{footnote}{1}
\footnotetext{Work done during an internship at FAIR.}
\renewcommand*{\thefootnote}{\arabic{footnote}}
\setcounter{footnote}{0}
\thispagestyle{empty}

\begin{abstract}
\vspace{-0.5em}
Most methods for object instance segmentation require all training examples to be labeled with segmentation masks. This requirement makes it expensive to annotate new categories and has restricted instance segmentation models to \app 100 well-annotated classes. The goal of this paper is to propose a new partially supervised training paradigm, together with a novel weight transfer function, that enables training instance segmentation models on a large set of categories all of which have box annotations, but only a small fraction of which have mask annotations. These contributions allow us to train Mask R-CNN to detect and segment 3000 visual concepts using box annotations from the Visual Genome dataset and mask annotations from the 80 classes in the COCO dataset. We evaluate our approach in a controlled study on the COCO dataset. This work is a first step towards instance segmentation models that have broad comprehension of the visual world. 
\end{abstract}
\vspace{-1em}

\secvspace
\section{Introduction}
\secvspace

Object detectors have become significantly more accurate (\eg, \cite{Girshick2014,Ren2015a}) and gained important new capabilities. One of the most exciting is the ability to predict a foreground segmentation mask for each detected object (\eg, \cite{He2017}), a task called \emph{instance segmentation}.
In practice, typical instance segmentation systems are restricted to a narrow slice of the vast visual world that includes only around 100 object categories.

A principle reason for this limitation is that state-of-the-art instance segmentation algorithms require \emph{strong supervision} and such supervision may be limited and expensive to collect for new categories \cite{Lin2014}. By comparison, bounding box annotations are more abundant and less expensive \cite{dai2015boxsup}. This fact raises a question: Is it possible to train high-quality instance segmentation models without complete instance segmentation annotations for all categories? With this motivation, our paper introduces a new \emph{partially supervised} instance segmentation task and proposes a novel transfer learning method to address it.

\begin{figure}[t]
\begin{center}
\vspace{-1em}
\includegraphics[width=0.9\linewidth]{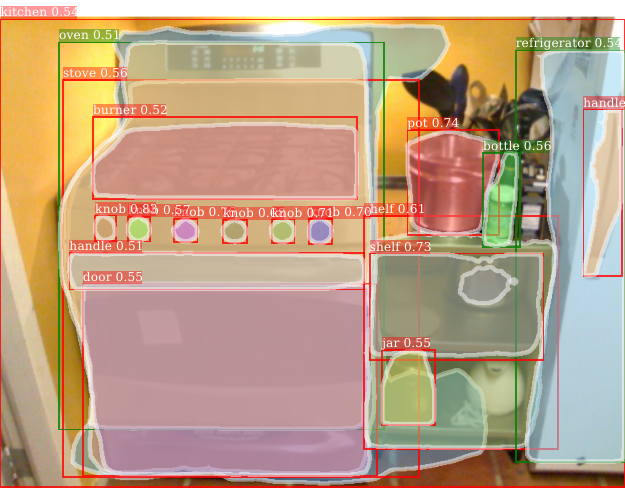}
\end{center}
\vspace{-1em}
\caption{\textbf{We explore training instance segmentation models with partial supervision}: a subset of classes (\textcolor{green}{green} boxes) have instance mask annotations during training; the remaining classes (\textcolor{red}{red} boxes) have only bounding box annotations. This image shows output from our model trained for 3000 classes from Visual Genome, using mask annotations from only 80 classes in COCO.}
\label{fig:teaser}
\vspace{-1.5em}
\end{figure}

We formulate the partially supervised instance segmentation task as follows: (1) given a set of categories of interest, a small \emph{subset} has instance mask annotations, while the other categories have only bounding box annotations; (2) the instance segmentation algorithm should utilize this data to fit a model that can segment instances of \emph{all} object categories in the set of interest. Since the training data is a mixture of strongly annotated examples (those with masks) and weakly annotated examples (those with only boxes), we refer to the task as \emph{partially supervised}.

The main benefit of partially supervised \vs weakly-supervised training (\cf \cite{khoreva2017simple}) is it allows us to build a large-scale instance segmentation model by exploiting both types of existing datasets: those with bounding box annotations over a large number of classes, such as Visual Genome \cite{krishna2017visual}, and those with instance mask annotations over a small number of classes, such as COCO \cite{Lin2014}. As we will show, this enables us to scale state-of-the-art instance segmentation methods to thousands of categories, a capability that is critical for their deployment in real world uses.

To address partially supervised instance segmentation, we propose a novel \emph{transfer learning} approach built on Mask R-CNN \cite{He2017}. Mask R-CNN is well-suited to our task because it decomposes the instance segmentation problem into the subtasks of bounding box object detection and mask prediction. These subtasks are handled by dedicated network `heads' that are trained jointly. The intuition behind our approach is that once trained, the parameters of the bounding box head encode an embedding of each object category that enables the transfer of visual information for that category to the partially supervised mask head.

We materialize this intuition by designing a parameterized \emph{weight transfer function} that is trained to predict a category's instance segmentation parameters as a function of its bounding box detection parameters. The weight transfer function can be trained end-to-end in Mask R-CNN using classes with mask annotations as supervision. At inference time, the weight transfer function is used to predict the instance segmentation parameters for \emph{every} category, thus enabling the model to segment all object categories, including those without mask annotations at training time.

We explore our approach in two settings. First, we use the COCO dataset \cite{Lin2014} to simulate the partially supervised instance segmentation task as a means of establishing quantitative results on a dataset with high-quality annotations and evaluation metrics. Specifically, we split the full set of COCO categories into a subset with mask annotations and a complementary subset for which the system has access to only bounding box annotations. Because the COCO dataset involves only a small number (80) of semantically well-separated classes, quantitative evaluation is precise and reliable. Experimental results show that our method improves results over a strong baseline with up to a 40\% relative increase in mask AP on categories without training masks.

In our second setting, we train a \emph{large-scale} instance segmentation model on 3000 categories using the Visual Genome (VG) dataset \cite{krishna2017visual}. VG contains bounding box annotations for a large number of object categories, however quantitative evaluation is challenging as many categories are semantically overlapping (\eg, near synonyms) and the annotations are not exhaustive, making precision and recall difficult to measure. Moreover, VG is not annotated with instance masks. Instead, we use VG to provide \emph{qualitative} output of a large-scale instance segmentation model. Output of our model is illustrated in Figure \ref{fig:teaser} and \ref{fig:vis_vg}.

\secvspace
\section{Related Work}
\secvspace

\begin{figure*}[t]
\begin{center}
\vspace{-2em}
\includegraphics[width=.98\linewidth]{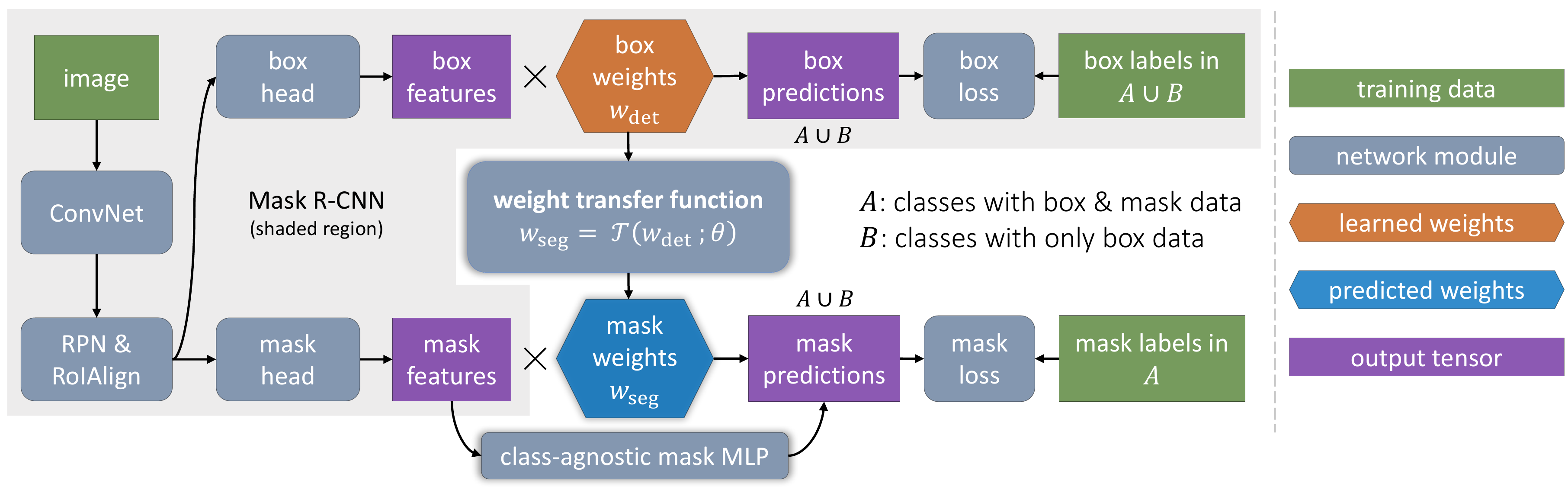}
\end{center}
\vspace{-1em}
\caption{\textbf{Detailed illustration of our \methodname method.} Instead of directly learning the mask prediction parameters $\wseg$, \methodname predicts a category's segmentation parameters $\wseg$ from its corresponding box detection parameters $\wdet$, using a learned weight transfer function $\ftrans$. For training, $\ftrans$ only needs mask data for the classes in set $A$, yet it can be applied to all classes in set $A\cup B$ at test time. We also augment the mask head with a complementary fully connected multi-layer perceptron (MLP).}
\label{fig:method}
\vspace{-1em}
\end{figure*}

\myparagraph{Instance segmentation.}
Instance segmentation is a highly active research area \cite{Hariharan2014,Hariharan2015,Dai2015, Pinheiro2015,Pinheiro2016,Dai2016,Hayder2017,Li2017,Kirillov2017,Bai2017}, with Mask R-CNN \cite{He2017} representing the current state-of-the-art. These methods assume a fully supervised training scenario in which \emph{all} categories of interest have instance mask annotations during training. Fully supervised training, however, makes it difficult to scale these systems to thousands of categories. The focus of our work is to relax this assumption and enable training models even when masks are available for only a small subset of categories. To do this, we develop a novel transfer learning approach built on Mask R-CNN.

\myparagraph{Weight prediction and task transfer learning.}
Instead of directly learning model parameters, prior work has explored \emph{predicting} them from other sources (\eg, \cite{ha2016hypernetworks}). In \cite{elhoseiny2013write}, image classifiers are predicted from the natural language description of a zero-shot category. In \cite{wang2016learning}, a model regression network is used to construct the classifier weights from few-shot examples, and similarly in \cite{misra2017red}, a small neural network is used to predict the classifier weights of the composition of two concepts from the classifier weights of each individual concept. Here, we design a model that predicts the class-specific instance segmentation weights used in Mask R-CNN, instead of training them directly, which is not possible in our partially supervised training scenario.

Our approach is also a type of transfer learning \cite{Pan2010} where knowledge gained from one task helps with another task. Most related to our work, LSDA \cite{hoffman2014lsda} transforms whole-image classification parameters into object detection parameters through a domain adaptation procedure. LSDA can be seen as transferring knowledge learned on an image classification task to an object detection task, whereas we consider transferring knowledge learned from bounding box detection to instance segmentation.

\myparagraph{Weakly supervised semantic segmentation.}
Prior work trains semantic segmentation models from weak supervision. Image-level labels and object size constraints are used in \cite{pathak2015constrained}, while other methods use boxes as supervision for expectation-maximization \cite{papandreou2015weakly} or iterating between proposals generation and training \cite{dai2015boxsup}. Point supervision and objectness potentials are used in \cite{bearman2016s}. Most work in this area addresses only semantic segmentation (not \emph{instance} segmentation), treats each class independently, and relies on hand-crafted bottom-up proposals that generalize poorly.

Weakly supervised instance segmentation is addressed in \cite{khoreva2017simple} by training an instance segmentation model over the bottom-up GrabCut \cite{rother2004grabcut} foreground segmentation results from the bounding boxes. Unlike \cite{khoreva2017simple}, we aim to exploit all existing labeled data rather than artificially limiting it. Our work is also complementary in the sense that bottom-up segmentation methods may be used to infer training masks for our weakly-labeled examples. We leave this extension to future work.

\myparagraph{Visual embeddings.}
Object categories may be modeled by continuous `embedding' vectors in a visual-semantic space, where nearby vectors are often close in appearance or semantic ontology. Class embedding vectors may be obtained via natural language processing techniques (\eg word2vec \cite{mikolov2013efficient} and GloVe \cite{pennington2014glove}), from visual appearance information (\eg \cite{dumoulin2016learned}), or both (\eg \cite{tsai2017learning}). In our work, the parameters of Mask R-CNN's box head contain class-specific appearance information and can be seen as embedding vectors learned by training for the bounding box object detection task. The class embedding vectors enable transfer learning in our model by sharing appearance information between visually related classes. We also compare with the NLP-based GloVe embeddings \cite{pennington2014glove} in our experiments.

\secvspace
\section{Learning to Segment Every Thing}
\label{sec:method}
\secvspace

Let $C$ be the set of object categories (\ie, `things' \cite{adelson2001seeing}) for which we would like to train an instance segmentation model. Most existing approaches assume that \emph{all} training examples in $C$ are annotated with instance masks. We relax this requirement and instead assume that $C = A \cup B$ where examples from the categories in $A$ have masks, while those in $B$ have only bounding boxes. Since the examples of the $B$ categories are weakly labeled \wrt the target task (instance segmentation), we refer to training on the combination of strong and weak labels as a \emph{partially supervised} learning problem. Noting that one can easily convert instance masks to bounding boxes, we assume that bounding box annotations are also available for classes in $A$.

Given an instance segmentation model like Mask R-CNN that has a bounding box detection component and a mask prediction component, we propose the \textbf{\methodname} method that transfers category-specific information from the model's bounding box detectors to its instance mask predictors.

\subsecvspace
\subsection{Mask Prediction Using Weight Transfer}
\label{sec:method_class_aware}
\subsecvspace

Our method is built on Mask R-CNN \cite{He2017}, because it is a simple instance segmentation model that also achieves state-of-the-art results. In brief, Mask R-CNN can be seen as augmenting a Faster R-CNN \cite{Ren2015a} bounding box detection model with an additional mask branch that is a small fully convolutional network (FCN) \cite{Long2015}. At inference time, the mask branch is applied to each detected object in order to predict an instance-level foreground segmentation mask. During training, the mask branch is trained jointly and in parallel with the standard bounding box head found in Faster R-CNN.

In Mask R-CNN, the last layer in the bounding box branch and the last layer in the mask branch both contain \emph{category-specific} parameters that are used to perform bounding box classification and instance mask prediction, respectively, for each category. Instead of learning the category-specific bounding box parameters and mask parameters independently, we propose to predict a category's mask parameters from its bounding box parameters using a generic, category-agnostic weight transfer function that can be jointly trained as part of the whole model.

For a given category $c$, let $\wcdet$ be the class-specific object detection weights in the last layer of the bounding box head, and $\wcseg$ be the class-specific mask weights in the mask branch. Instead of treating $\wcseg$ as model parameters, $\wcseg$ is parameterized using a generic weight prediction function $\ftrans(\cdot)$:
\begin{equation}
\wcseg = \ftrans(\wcdet; \thetatrans),
\end{equation}
where $\thetatrans$ are class-agnostic, learned parameters.

The \emph{same} transfer function $\ftrans(\cdot)$ may be applied to any category $c$ and, thus, $\thetatrans$ should be set such that $\ftrans$ generalizes to classes whose masks are not observed during training. We expect that generalization is possible because the class-specific detection weights $\wcdet$ can be seen as an appearance-based visual embedding of the class.

$\ftrans(\cdot)$ can be implemented as a small fully connected neural network. Figure \ref{fig:method} illustrates how the weight transfer function fits into Mask R-CNN to form \methodname. As a detail, note that the bounding box head contains two types of detection weights: the RoI classification weights $\wccls$ and the bounding box regression weights $\wcbox$. We experiment with using either only a single type of detection weights (\ie $\wcdet=\wccls$ or $\wcdet=\wcbox$) or using the concatenation of the two types of weights (\ie $\wcdet=[\wccls, \wcbox]$).

\subsecvspace
\subsection{Training}
\label{sec:method_training}
\subsecvspace

During training, we assume that for the two sets of classes $A$ and $B$, instance mask annotations are available only for classes in $A$ but not for classes in $B$, while all classes in $A$ and $B$ have bounding box annotations available. As shown in Figure \ref{fig:method}, we train the bounding box head using the standard box detection losses on all classes in $A \cup B$, but only train the mask head and the weight transfer function $\ftrans(\cdot)$ using a mask loss on the classes in $A$. Given these losses, we explore two different training procedures: stage-wise training and end-to-end training.

\myparagraph{Stage-wise training.}
As Mask R-CNN can be seen as augmenting Faster R-CNN with a mask head, a possible training strategy is to separate the training procedure into detection training (first stage) and segmentation training (second stage). In the first stage, we train a Faster R-CNN using only the bounding box annotations of the classes in $A \cup B$, and then in the second stage the additional mask head is trained while keeping the convolutional features and the bounding box head fixed. In this way, the class-specific detection weights $\wcdet$ of each class $c$ can be treated as fixed class embedding vectors that do not need to be updated when training the second stage. This approach has the practical benefit of allowing us to train the box detection model once and then rapidly evaluate design choices for the weight transfer function. It also has disadvantages, which we discuss next.

\myparagraph{End-to-end joint training.}
It was shown that for Mask R-CNN, multi-task training can lead to better performance than training on each task separately. The aforementioned stage-wise training mechanism separates detection training and segmentation training, and may result in inferior performance. Therefore, we would also like to jointly train the bounding box head and the mask head in an end-to-end manner. In principle, one can directly train with back-propagation using the box losses on classes in $A \cup B$ and the mask loss on classes in $A$. However, this may cause a discrepancy in the class-specific detection weights $\wcdet$ between set $A$ and $B$, since only $\wcdet$ for $c \in A$ will receive gradients from the mask loss through the weight transfer function $\ftrans(\cdot)$. We would like $\wcdet$ to be homogeneous between $A$ and $B$ so that the predicted $\wcseg = \ftrans(\wcdet; \thetatrans)$ trained on $A$ can better generalize to $B$.

To address this discrepancy, we take a simple approach: when back-propagating the mask loss, we stop the gradient with respect to $\wcdet$, that is, we only compute the gradient of the predicted mask weights $\ftrans(\wcdet; \thetatrans)$ with respect to transfer function parameter $\thetatrans$ but not bounding box weight $\wcdet$. This can be implemented as $\wcseg = \ftrans(\mathtt{stop\_grad}(\wcdet); \thetatrans)$ in most neural network toolkits.

\subsecvspace
\subsection{Baseline: Class-Agnostic Mask Prediction}
\label{sec:method_class_agnostic}
\subsecvspace

DeepMask \cite{Pinheiro2015} established that it is possible to train a deep model to perform \emph{class-agnostic} mask prediction where an object mask is predicted regardless of the category. A similar result was also shown for Mask R-CNN with only a small loss in mask quality \cite{He2017}. In additional experiments, \cite{Pinheiro2015} demonstrated if the class-agnostic model is trained to predict masks on a \emph{subset} of the COCO categories (specifically the 20 from PASCAL VOC \cite{Everingham2010}) it can \emph{generalize} to the other 60 COCO categories at inference time. Based on these results, we use Mask R-CNN with a class-agnostic FCN mask prediction head as a baseline. Evidence from \cite{Pinheiro2015} and \cite{He2017} suggest that this is a strong baseline. Next, we introduce an extension that can improve both the baseline and our proposed weight transfer function.

We also compare with a few other baselines for unsupervised or weakly supervised instance segmentation in \S\ref{sec:final_comp}.

\subsecvspace
\subsection{Extension: Fused FCN+MLP Mask Heads}
\label{sec:method_mlp}
\subsecvspace

Two types of mask heads are considered for Mask R-CNN in \cite{He2017}: (1) an FCN head, where the $M \times M$ mask is predicted with a fully convolutional network, and (2) an MLP head, where the mask is predicted with a multi-layer perceptron consisting of fully connected layers, more similar to DeepMask. In Mask R-CNN, the FCN head yields higher mask AP. However, the two designs may be complementary. Intuitively, the MLP mask predictor may better capture the `gist' of an object while the FCN mask predictor may better capture the details (such as the object boundary). Based on this observation, we propose to improve both the baseline class-agnostic FCN and our weight transfer function (which uses an FCN) by fusing them with predictions from a class-agnostic MLP mask predictor. Our experiments will show that this extension brings improvements to both the baseline and our transfer approach.

When fusing class-agnostic and class-specific mask predictions for $K$ classes, the two scores are added into a final $K \times M \times M$ output, where the class-agnostic mask scores (with shape $1 \times M \times M$) are tiled $K$ times and added to every class. Then, the $K \times M \times M$ mask scores are turned into per-class mask probabilities through a sigmoid unit, and resized to the actual bounding box size as final instance mask for that bounding box. During training, binary cross-entropy loss is applied on the $K \times M \times M$ mask probabilities.

\secvspace
\section{Experiments on COCO}
\secvspace

We evaluate our method on the COCO dataset \cite{Lin2014}, which is small scale \wrt the number of categories but contains exhaustive mask annotations for 80 categories. This property enables rigorous quantitative evaluation using standard detection metrics, like average precision (AP).

\subsecvspace
\subsection{Evaluation Protocol and Baselines}
\label{sec:exp_coco}
\subsecvspace

\begin{table*}
\vspace{-3em}
\begin{center}
\subfloat[\textbf{Ablation on input to $\ftrans$.} `cls' is RoI classification weights, `box' is box regression weights, and`cls+box' is both weights. We also compare with the NLP-based GloVe vectors \cite{pennington2014glove}. Our transfer function $\ftrans$ improves the AP on $B$ while remaining on par with the oracle on $A$. \label{tab:result_coco_ablation_input}]{\tablestyle{2.5pt}{1.05}
\begin{tabular}{l|x{29}x{29}|x{29}x{29}}
& \multicolumn{2}{c|}{voc $\rightarrow$ non-voc} & \multicolumn{2}{c}{non-voc $\rightarrow$ voc} \\
method & AP on $B$ & AP on $A$ &  AP on $B$ & AP on $A$ \\
\shline
transfer w/ randn & 15.4 & \demph{35.2} & 19.9 & \demph{31.1} \\
transfer w/ GloVe \cite{pennington2014glove} & 17.3 & \demph{35.2} & 21.9 & \demph{31.1} \\
transfer w/ cls &	18.1 & \demph{35.1} & 25.2 & \demph{31.1} \\
transfer w/ box & 19.8 & \demph{35.2} & 25.7 & \demph{31.1} \\
transfer w/ cls+box &	\textbf{20.2} & \demph{35.2} & \textbf{26.0} & \demph{31.2} \\
\hline
class-agnostic (baseline) &14.2 & \demph{34.4} & 21.5 & \demph{30.7} \\
fully supervised (oracle) & 30.7 & \demph{35.0} & 35.0 & \demph{30.7} \\
\end{tabular}
}
~~~~
\subfloat[\textbf{Ablation on the structure of $\ftrans$.} We vary the number of fully connected layers in the weight transfer function $\ftrans$, and try both ReLU and LeakyReLU as activation function in the hidden layers. The results show that `2-layer, LeakyReLU' works best, but in general $\ftrans$ is robust to specific implementation choices. \label{tab:result_coco_ablation_structure}]{\tablestyle{2.5pt}{1.05}
\begin{tabular}{p{110pt}|x{29}x{29}|x{29}x{29}}
& \multicolumn{2}{c|}{voc $\rightarrow$ non-voc} & \multicolumn{2}{c}{non-voc $\rightarrow$ voc} \\
method & AP on $B$ & AP on $A$ &  AP on $B$ & AP on $A$ \\
\shline
transfer w/ 1-layer, none & 19.2 & \demph{35.2} & 25.3 & \demph{31.2} \\
transfer w/ 2-layer, ReLU & 19.7 & \demph{35.3} & 25.1 & \demph{31.1} \\
transfer w/ 2-layer, LeakyReLU & \textbf{20.2} & \demph{35.2} & \textbf{26.0} & \demph{31.2} \\
transfer w/ 3-layer, ReLU & 19.3 & \demph{35.2} & \textbf{26.0} & \demph{31.1} \\
transfer w/ 3-layer, LeakyReLU & 18.9 & \demph{35.2} & 25.5 & \demph{31.1} \\
\multicolumn{5}{c}{} \\
\multicolumn{5}{c}{}
\end{tabular}
}
\vspace{-1em}
\subfloat[\textbf{Impact of the MLP mask branch.} Adding the class-agnostic MLP mask branch (see \S\ref{sec:method_mlp}) improves the performance of classes in set $B$ for both the class-agnostic baseline and our weight transfer approach. \label{tab:result_coco_ablation_mlp}]{\tablestyle{2.5pt}{1.05}
\begin{tabular}{l|x{29}x{29}|x{29}x{29}}
& \multicolumn{2}{c|}{voc $\rightarrow$ non-voc} & \multicolumn{2}{c}{non-voc $\rightarrow$ voc} \\
method & AP on $B$ & AP on $A$ &  AP on $B$ & AP on $A$ \\
\shline
class-agnostic~~~~~~~~~~~~~~~~~~& 14.2 & \demph{34.4} & 21.5 & \demph{30.7} \\
class-agnostic+MLP & 17.1 & \demph{35.1} & 22.8 & \demph{31.3} \\
transfer & 20.2 & \demph{35.2} & 26.0 & \demph{31.2} \\
transfer+MLP & \textbf{21.3} & \demph{35.4} & \textbf{26.6} & \demph{31.4} \\
\multicolumn{5}{c}{}
\end{tabular}
}
~~~~
\subfloat[\textbf{Ablation on the training strategy.} We try both stage-wise (`sw') and end-to-end (`e2e') training (see \S\ref{sec:method_training}), and whether to stop gradient from $\ftrans$ to $\wdet$. End-to-end training improves the results and it is crucial to stop gradient on $\wdet$.\label{tab:result_coco_ablation_training}]{\tablestyle{2.5pt}{1.05}
\begin{tabular}{lx{25}x{30}|x{29}x{29}|x{29}x{29}}
& & stop grad & \multicolumn{2}{c|}{voc $\rightarrow$ non-voc} & \multicolumn{2}{c}{non-voc $\rightarrow$ voc} \\
method & training & on $\wdet$ & AP on $B$ & AP on $A$ &  AP on $B$ & AP on $A$ \\
\shline
class-agnostic & sw & \demph{n/a} & 14.2 & \demph{34.4} & 21.5 & \demph{30.7} \\
transfer & sw & \demph{n/a} & 20.2 & \demph{35.2} & 26.0 & \demph{31.2} \\
\hline
class-agnostic & e2e & \demph{n/a} & 19.2 & \demph{36.8} & 23.9 & \demph{32.5} \\
transfer & e2e &  & 20.2 & \demph{37.7} & 24.8 & \demph{33.2} \\
transfer & e2e & \checkmark & \textbf{22.2} & \demph{37.6} & \textbf{27.6} & \demph{33.1} \\
\end{tabular}
}
\end{center}
\vspace{-0.5em}
\caption{\textbf{Ablation study of our method.} We use ResNet-50-FPN as our backbone network, and `cls+box' and `2-layer, LeakyReLU' as the default input and structure of $\ftrans$. Results in (a,b,c) are based on stage-wise training, and we study the impact of end-to-end training in (d). Mask AP is evaluated on the COCO dataset \texttt{val2017} split between the 20 PASCAL VOC categories (`voc') and the 60 remaining categories (`non-voc'), as in \cite{Pinheiro2015}. Performance on the strongly supervised set $A$ is shown in \demph{gray}.}
\label{tab:result_coco_ablation}
\vspace{-1em}
\end{table*}

We simulate the partially supervised training scenario on COCO by partitioning the 80 classes into sets $A$ and $B$, as described in \S\ref{sec:method}. We consider two split types: (1) The 20/60 split used by DeepMask \cite{Pinheiro2015} that divides the COCO categories based on the 20 contained in PASCAL VOC \cite{Everingham2010} and the 60 that are not. We refer to these as the `voc' and `non-voc' category sets from here on. (2) We also conduct experiments using multiple trials with random splits of different sizes. These experiments allow us to characterize any bias in the voc/non-voc split and also understand what factors in the training data lead to better mask generalization.

\myparagraph{Implementation details.} We train our model on the COCO \texttt{train2017} split and test on \texttt{val2017}.\footnote{The COCO \texttt{train2017} and \texttt{val2017} splits are the same as the \texttt{trainval35k} and \texttt{minival} splits used in prior work, such as \cite{He2017}.} Each class has a 1024-d RoI classification parameter vector $\wccls$ and a 4096-d bounding box regression parameter vector $\wcbox$ in the detection head, and a 256-d segmentation parameter vector $\wcseg$ in the mask head. The output mask resolution is $M \times M = 28 \times 28$. In all our experimental analysis below, we use either ResNet-50-FPN or ResNet-101-FPN \cite{Lin2017} as the backbone architecture for Mask R-CNN, initialized from a ResNet-50 or a ResNet-101 \cite{He2016} model pretrained on the ImageNet-1k image classification dataset \cite{Russakovsky2015}.

We follow the training hyper-parameters suggested for Mask R-CNN in \cite{He2017}. Each minibatch has 16 images $\times$ 512 RoIs-per-images, and the network is trained for 90k iterations on 8 GPUs. We use 1e-4 weight decay and 0.9 momentum, and an initial learning rate of 0.02, which is multiplied by 0.1 after 60k and 80k iterations. We evaluate instance segmentation performance using average precision (AP), which is the standard COCO metric and equal to the mean of average precision from 0.5 to 0.95 IoU threshold of all classes.

\myparagraph{Baseline and oracle.}
We compare our method to class-agnostic mask prediction using either an FCN or fused FCN+MLP structure. In these approaches, instead of predicting each class $c$'s segmentation parameters $\wcseg$ from its bounding box classification parameters $\wcdet$, all the categories share the \emph{same} learned segmentation parameters $\wcseg$ (no weight transfer function is involved). Evidence from DeepMask and Mask R-CNN, as discussed in \S\ref{sec:method_class_agnostic}, suggests that this approach is a strong baseline.
In addition, we compare our approach with unsupervised or weakly supervised instance segmentation approaches in \S\ref{sec:final_comp}.

We also evaluate an `oracle' model: Mask R-CNN trained on all classes in $A \cup B$ with access to instance mask annotations for \emph{all} classes in $A$ and $B$ at training time. This fully supervised model is a performance upper bound for our partially supervised task (unless the weight transfer function can improve over directly learning $\wcseg$).

\subsecvspace
\subsection{Ablation Experiments}
\subsecvspace

\myparagraph{Input to $\ftrans$.}
In Table \ref{tab:result_coco_ablation_input} we study the impact of the input to the weight transfer function $\ftrans$. For transfer learning to work, we expect that the input should capture information about how the visual appearance of classes relate to each other. To see if this is the case, we designed several inputs to $\ftrans$: a random Gaussian vector (`randn') assigned to each class, an NLP-based word embedding using pre-trained GloVe vectors \cite{pennington2014glove} for each class, the weights from the Mask R-CNN box head classifier (`cls'), the weights from the box regression (`box'), and the concatenation of both weights (`cls+box'). We compare the performance of our transfer approach with these different embeddings to the strong baseline: class-agnostic Mask R-CNN.

\begin{figure}[t]
\begin{center}
\vspace{-1.5em}
\includegraphics[width=0.95\linewidth]{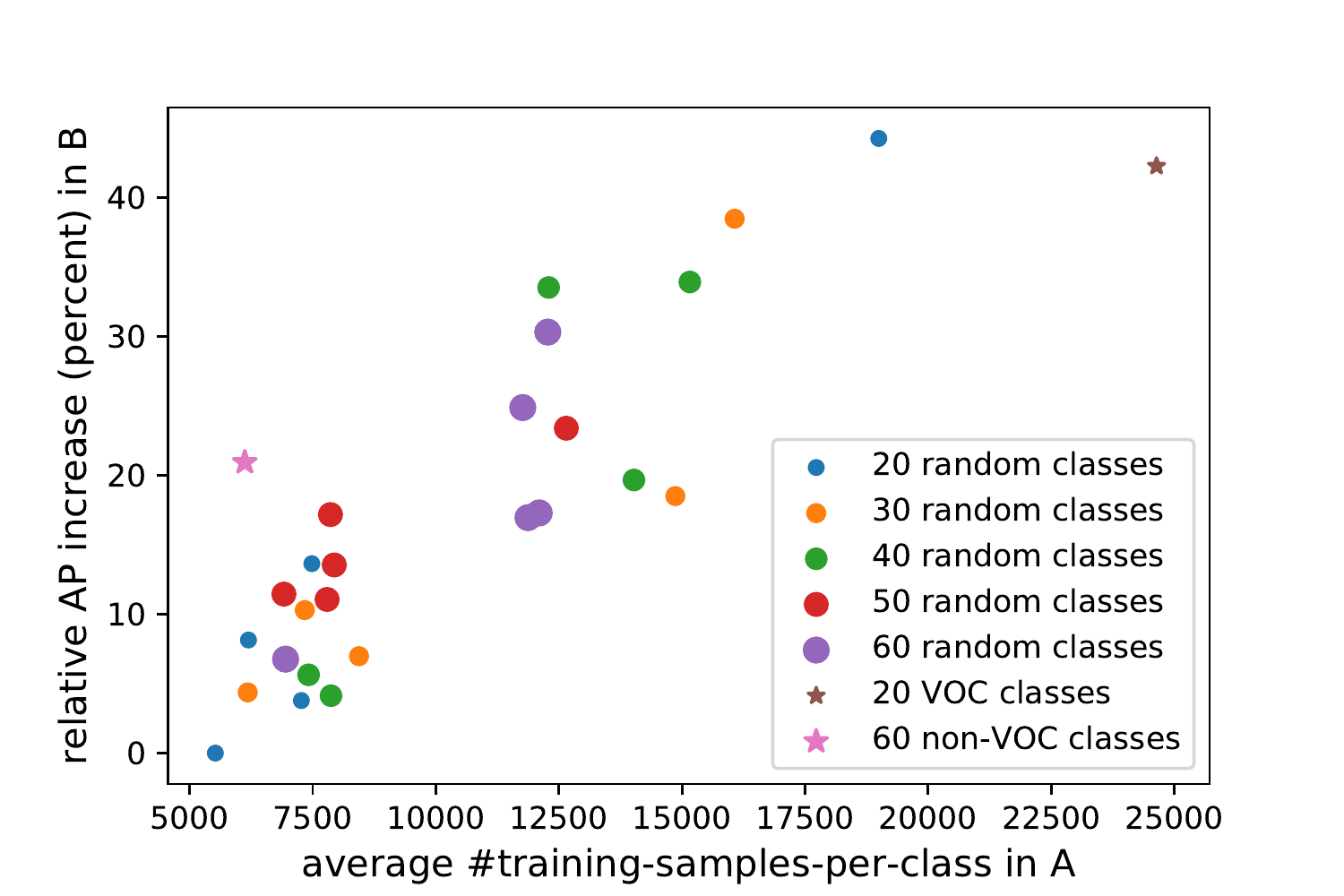} \\
\vspace{-1.5em}
\end{center}
\caption{Each point corresponds to our method on a random $A$/$B$ split of COCO classes. We vary $|A|$ from 20 to 60 classes and plot the relative change in mask AP on the classes in set $B$ (those classes without mask annotations) \vs the average number of mask annotations per class in set $A$.}
\label{fig:exp_ablation_rand_split}
\vspace{-1.5em}
\end{figure}

First, Table \ref{tab:result_coco_ablation_input} shows that the random control (`randn') yields results on par with the baseline; they are slightly better on voc$\rightarrow$non-voc and worse in the other direction, which may be attributed to noise. Next, the GloVe embedding shows a consistent improvement over the baseline, which indicates that these embeddings may capture some visual information as suggested in prior work \cite{xian2016latent}. However, inputs `cls', `box' and `cls+box' all strongly outperform the NLP-based embedding (with `cls+box' giving the best results), which matches our expectation since they encode visual information by design.

We note that all methods compare well to the fully supervised Mask R-CNN oracle on the classes in set $A$. In particular, our transfer approach slightly outperforms the oracle for all input types. This results indicates that our approach does not sacrifice anything on classes with strong supervision, which is an important property.

\myparagraph{Structure of $\ftrans$.}
In Table \ref{tab:result_coco_ablation_structure} we compare different implementations of $\ftrans$: as a simple affine transformation, or as a neural network with 2 or 3 layers. Since LeakyReLU \cite{maas2013rectifier} is used for weight prediction in \cite{misra2017red}, we try both ReLU and LeakyReLU as activation function in the hidden layers. The results show that a 2-layer MLP with LeakyReLU gives the best mask AP on set $B$. Given this, we select the `cls+box, 2-layer, LeakyReLU' implementation of $\ftrans$ for all subsequent experiments.

\myparagraph{Comparison of random $A$/$B$ splits.}
Besides splitting datasets into voc and non-voc, we also experiment with random splits of the 80 classes in COCO, and vary the number of training classes. We randomly select 20, 30, 40, 50 or 60 classes to include in set $A$ (the complement forms set $B$), perform 5 trials for each split size, and compare the performance of our weight transfer function $\ftrans$ on classes in $B$ to the class-agnostic baseline. The results are shown in Figure \ref{fig:exp_ablation_rand_split}, where it can be seen that our method yields to up to over 40\% relative increase in mask AP. This plot reveals a correlation between relative AP increase and the average number of training samples per class in set $A$. This indicates that to maximize transfer performance to classes in set $B$ it may be more effective to collect a larger number of instance mask samples for each object category in set $A$.

\begin{figure*}[!ht]
\begin{center}
\vspace{-1.5em}
\includegraphics[height=0.17\linewidth]{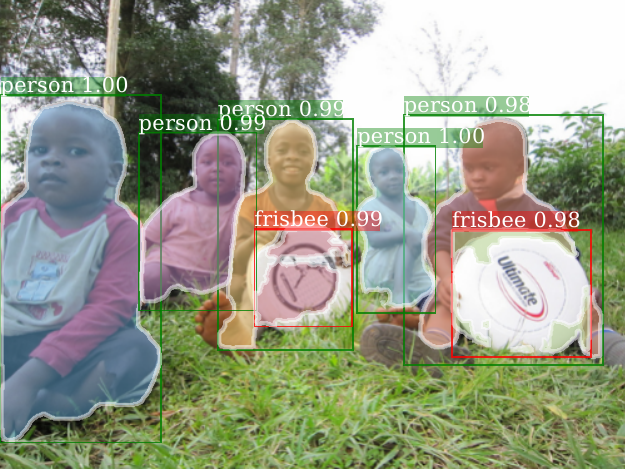}
\includegraphics[height=0.17\linewidth]{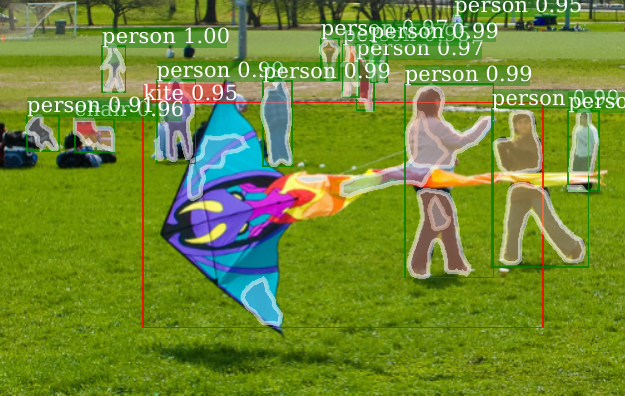}
\includegraphics[height=0.17\linewidth]{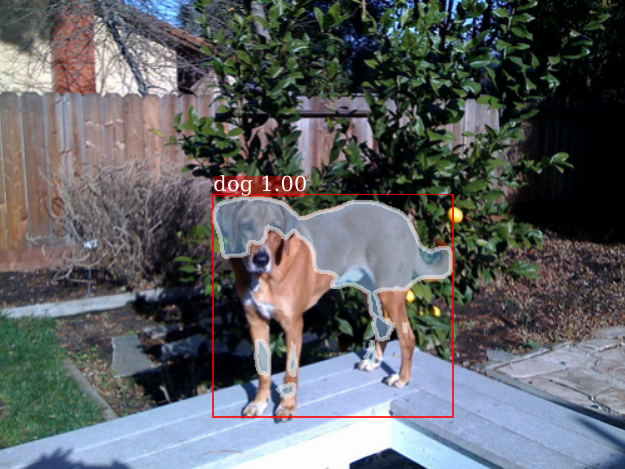}
\includegraphics[height=0.17\linewidth]{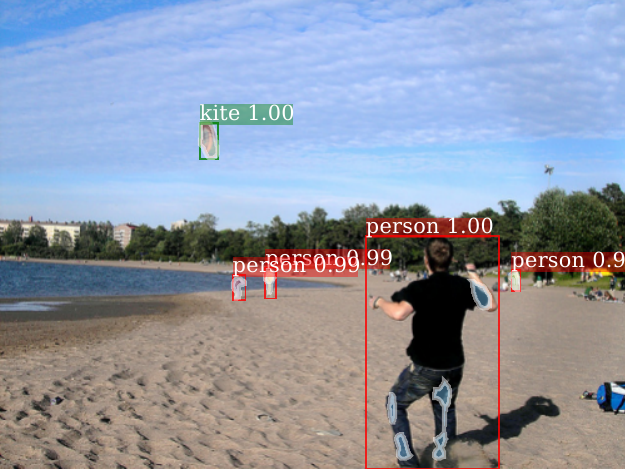} \\
\vspace{0.3mm}
\includegraphics[height=0.17\linewidth]{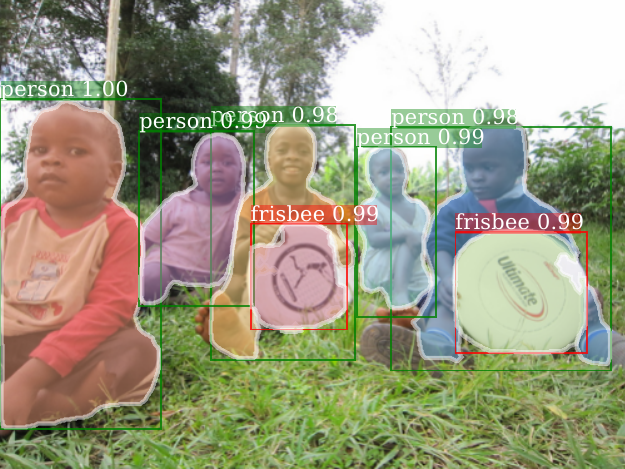}
\includegraphics[height=0.17\linewidth]{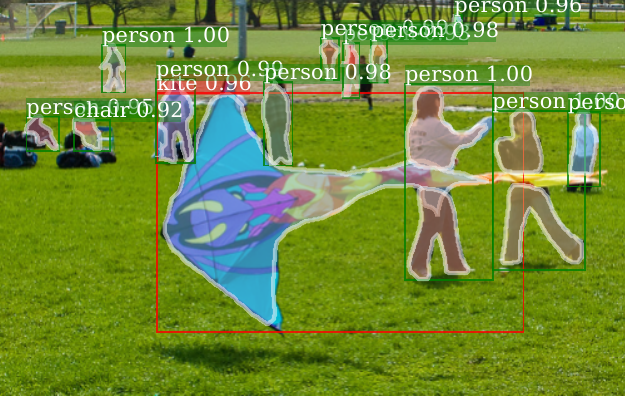}
\includegraphics[height=0.17\linewidth]{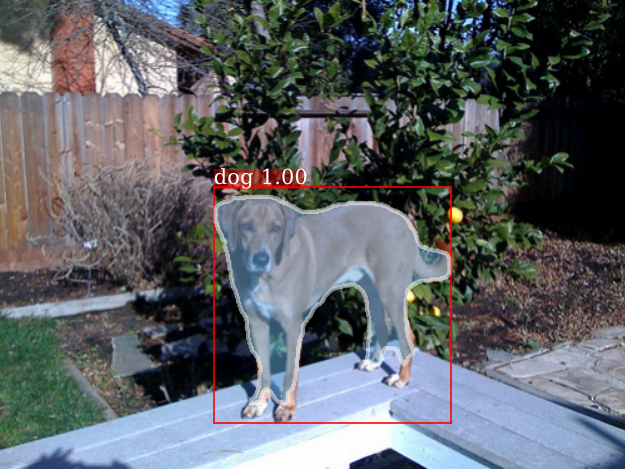}
\includegraphics[height=0.17\linewidth]{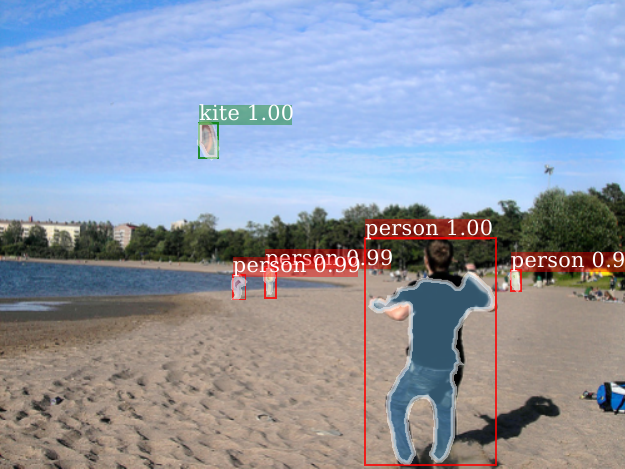} \\
\end{center}
\vspace{-1em}
\caption{\textbf{Mask predictions from the class-agnostic baseline (top row) \vs our \methodname approach (bottom row)}. \textcolor{green}{Green} boxes are classes in set $A$ while the \textcolor{red}{red} boxes are classes in set $B$. The left 2 columns are $A=\{\text{voc}\}$ and the right 2 columns are $A=\{\text{non-voc}\}$.}
\label{fig:vis_coco}
\vspace{-1em}
\end{figure*}
%
\begin{table*}
\tablestyle{3.5pt}{1.1}
\begin{center}
\begin{tabular}{ll|x{22}x{22}x{22}x{22}x{22}x{22}|x{22}x{22}x{22}x{22}x{22}x{22}}
& & \multicolumn{6}{c|}{voc $\rightarrow$ non-voc: test on $B=\{\text{non-voc}\}$} & \multicolumn{6}{c}{non-voc $\rightarrow$ voc: test on $B=\{\text{voc}\}$} \\
backbone & method & AP & AP$_{50}$ & AP$_{75}$ & AP$_S$ & AP$_M$ & AP$_L$ & AP & AP$_{50}$ & AP$_{75}$ & AP$_S$ & AP$_M$ & AP$_L$ \\
\shline
& class-agnostic &
19.2 & 36.4 & 18.4 & 11.5 & 23.3 & 24.4 & 23.9 & 42.9 & 23.5 & 11.6 & 24.3 & 33.7 \\
& Faster R-CNN tested w/ GrabCut &
12.6 & 24.0 & 11.9 &  4.3 & 12.0 & 23.5 & 12.1 & 27.7 &  8.9 & 4.3 & 12.0 & 23.5 \\
R-50-FPN & Mask R-CNN trained w/ GrabCut &
19.5 & 39.2 & 17.0 &  6.5 & 20.9 & \textbf{34.3} & 19.5 & 46.2 & 14.2 & 4.7 & 15.9 & 32.0 \\
& \methodname (ours) &
\textbf{23.7} & \textbf{43.1} & \textbf{23.5} & \textbf{12.4} & \textbf{27.6} & 32.9 & \textbf{28.9} & \textbf{52.2} & \textbf{28.6} & \textbf{12.1} & \textbf{29.0} & \textbf{40.6} \\
& fully supervised (oracle)
& \demph{33.0} & \demph{53.7} & \demph{35.0} & \demph{15.1} & \demph{37.0} & \demph{49.9}       & \demph{37.5} & \demph{63.1} & \demph{38.9} & \demph{15.1} & \demph{36.0} & \demph{53.1}
\\
\hline
& class-agnostic &
18.5 & 34.8 & 18.1 & 11.3 & 23.4 & 21.7 & 24.7 & 43.5 & 24.9 & 11.4 & 25.7 & 35.1 \\
& Faster R-CNN tested w/ GrabCut & 13.0 & 24.6 & 12.1 & 4.5 & 12.3 & 24.4 & 12.3 & 27.6 & 9.5 & 4.5 & 12.3 & 24.4 \\
R-101-FPN & Mask R-CNN trained w/ GrabCut & 19.7 & 39.7 & 17.0 & 6.4 & 21.2 & \textbf{35.8} & 19.6 & 46.1 & 14.3 & 5.1 & 16.0 & 32.4 \\
& \methodname (ours) &
\textbf{23.8} & \textbf{42.9} & \textbf{23.5} & \textbf{12.7} & \textbf{28.1} & 33.5 & \textbf{29.5} & \textbf{52.4} & \textbf{29.7} & \textbf{13.4} & \textbf{30.2} & \textbf{41.0} \\
& fully supervised (oracle)
& \demph{34.4} & \demph{55.2} & \demph{36.3} & \demph{15.5} & \demph{39.0} & \demph{52.6} & \demph{39.1} & \demph{64.5} & \demph{41.4} & \demph{16.3} & \demph{38.1} & \demph{55.1}
\\
\end{tabular}
\end{center}
\vspace{-1em}
\caption{\textbf{End-to-end training of \methodname.} As in Table \ref{tab:result_coco_ablation}, we use `cls+box, 2-layer, LeakyReLU' implementation of $\ftrans$ and add the MLP mask branch (`transfer+MLP'), and follow the same evaluation protocol.  We also report AP$_{50}$ and AP$_{75}$ (average precision evaluated at 0.5 and 0.75 IoU threshold respectively), and AP over small (AP$_S$), medium (AP$_M$), and large (AP$_{L}$) objects. Our method significantly outperforms the baseline approaches in \S\ref{sec:final_comp} on set $B$ without mask training data for both ResNet-50-FPN and ResNet-101-FPN backbones.}
\label{tab:result_coco_e2e}
\vspace{-1.5em}
\end{table*}

\myparagraph{Impact of the MLP mask branch.} 
As discussed in \S\ref{sec:method_mlp}, a class-agnostic MLP mask branch can be fused with either the baseline or our transfer approach. In Table \ref{tab:result_coco_ablation_mlp} we see that either mask head fused with the MLP mask branch consistently outperforms the corresponding unfused version. This confirms our intuition that FCN-based mask heads and MLP-based mask heads are complementary in nature.

\myparagraph{Effect of end-to-end training.}
Up to now, all ablation experiments use stage-wise training, because it is significantly faster (the same Faster R-CNN detection model can be reused for all experiments). However, as noted in \S\ref{sec:method_training}, stage-wise training may be suboptimal. Thus, Table \ref{tab:result_coco_ablation_training} compares stage-wise training to end-to-end training. In the case of end-to-end training, we investigate if it is necessary to stop gradients from $\ftrans$ to $\wdet$, as discussed. Indeed, results match our expectation that end-to-end training can bring improved results, \emph{however only when back-propagation from $\ftrans$ to $\wdet$ is disabled}. We believe this modification is necessary in order to make the embedding of classes in $A$ homogeneous with those in $B$; a property that is destroyed when only the embeddings for classes in $A$ are modified by back-propagation from $\ftrans$.

\begin{figure*}[t]
\begin{center}
\vspace{-2em}
\includegraphics[width=0.24\linewidth]{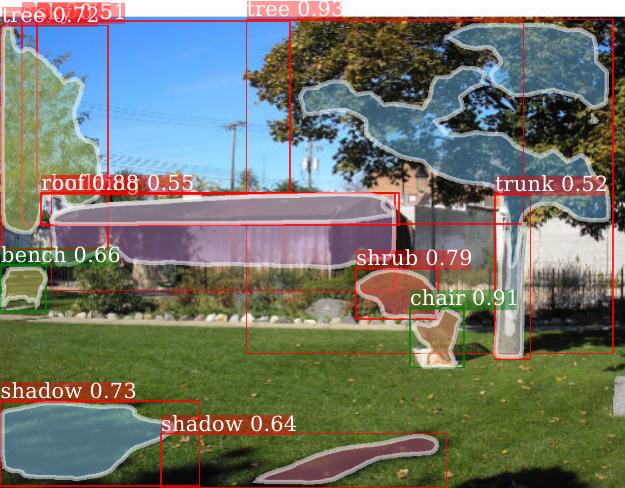}
\includegraphics[width=0.24\linewidth]{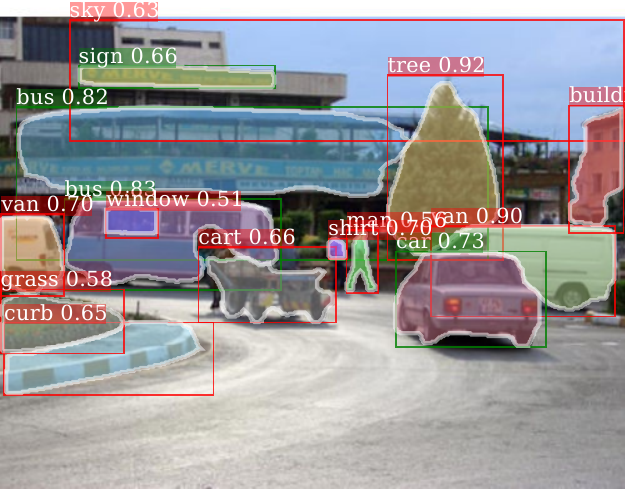}
\includegraphics[width=0.24\linewidth]{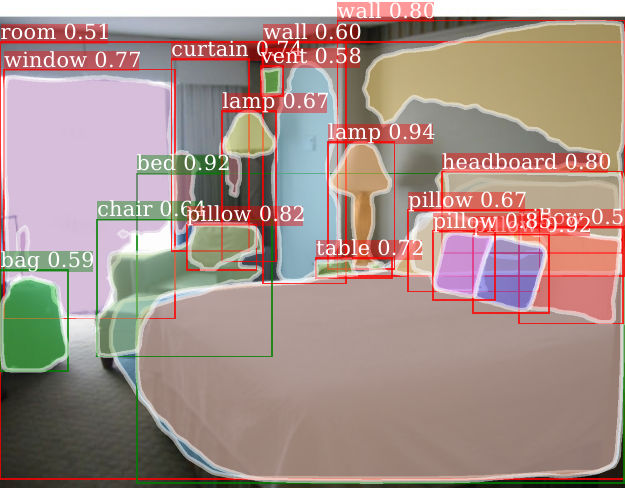}
\includegraphics[width=0.24\linewidth]{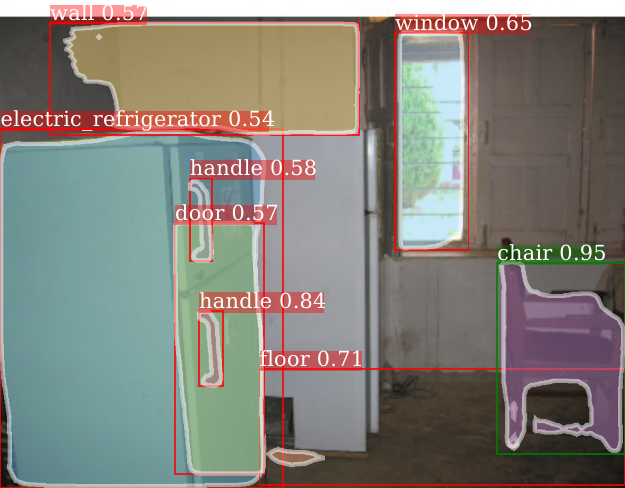} \\
\includegraphics[width=0.24\linewidth]{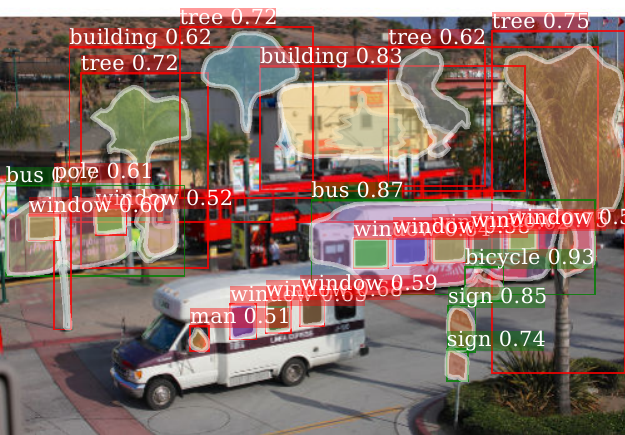}
\includegraphics[width=0.24\linewidth]{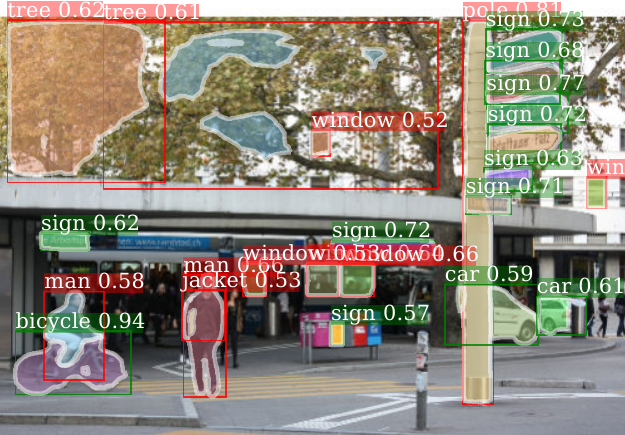}
\includegraphics[width=0.24\linewidth]{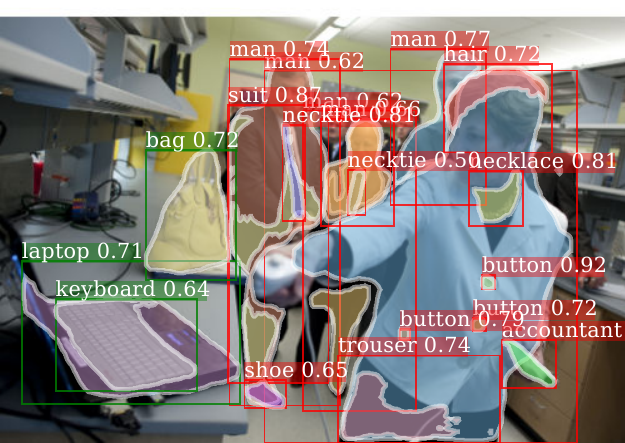}
\includegraphics[width=0.24\linewidth]{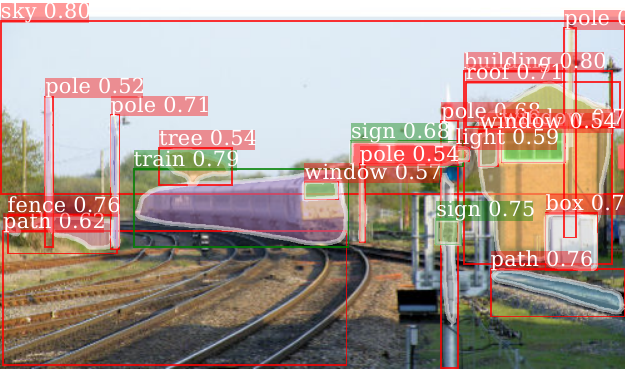} \\
\includegraphics[width=0.24\linewidth]{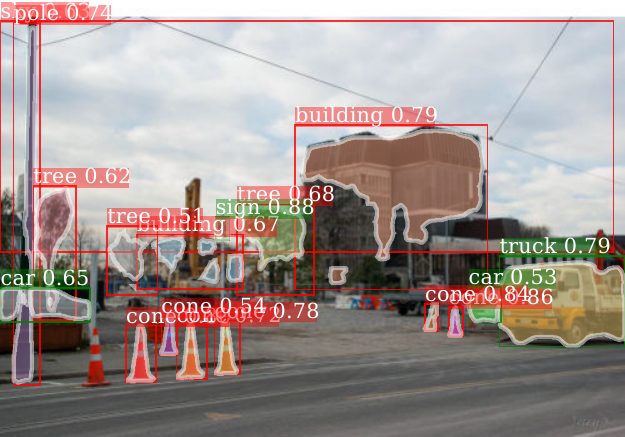}
\includegraphics[width=0.24\linewidth]{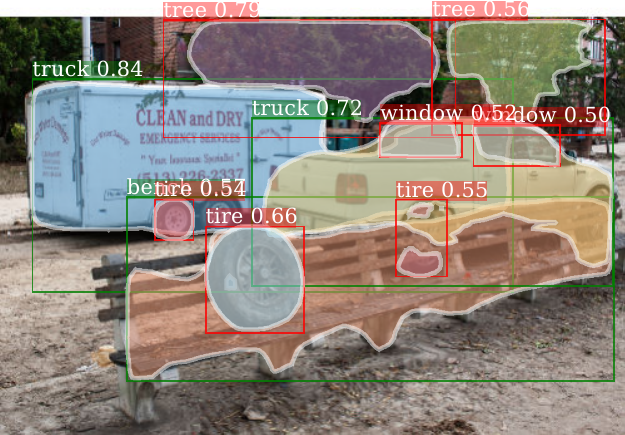}
\includegraphics[width=0.24\linewidth]{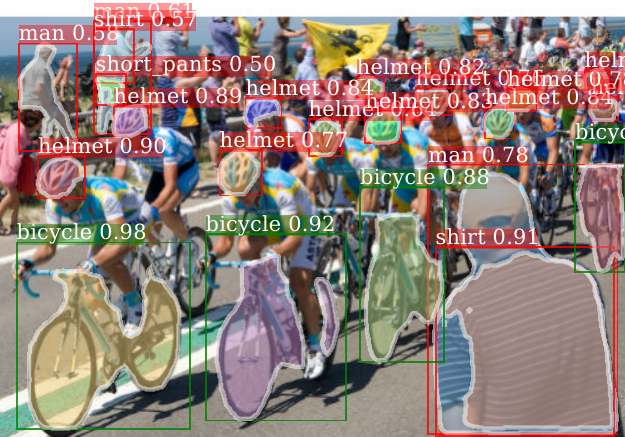}
\includegraphics[width=0.24\linewidth]{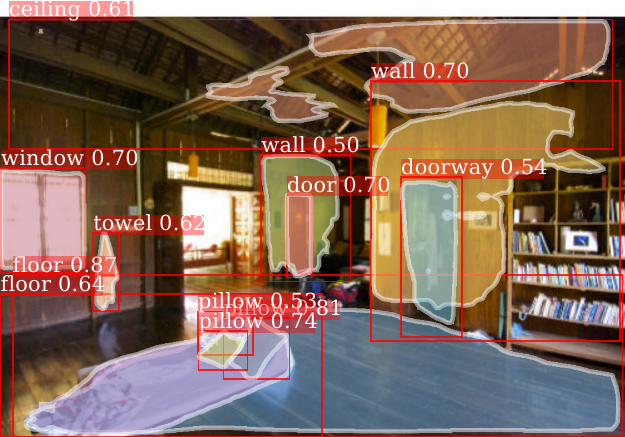} \\
\includegraphics[width=0.24\linewidth]{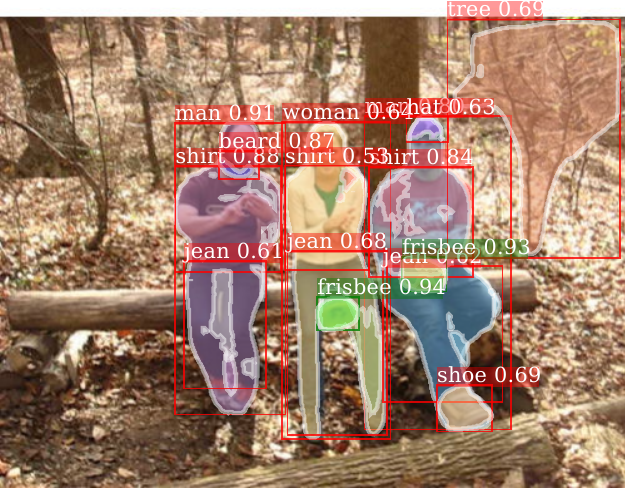}
\includegraphics[width=0.24\linewidth]{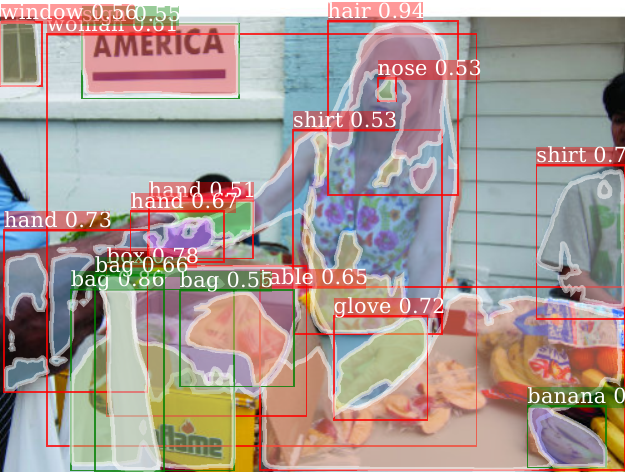}
\includegraphics[width=0.24\linewidth]{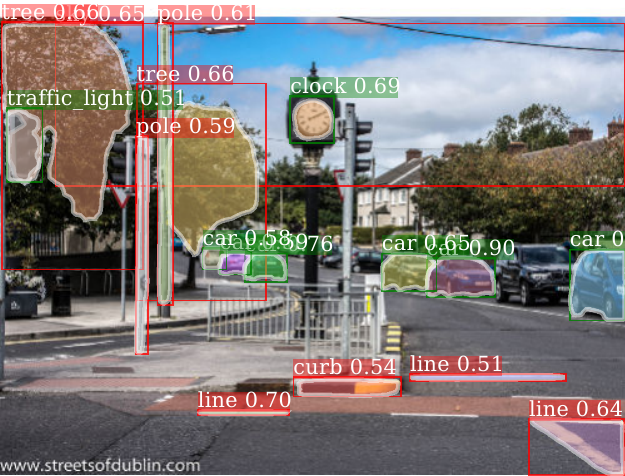}
\includegraphics[width=0.24\linewidth]{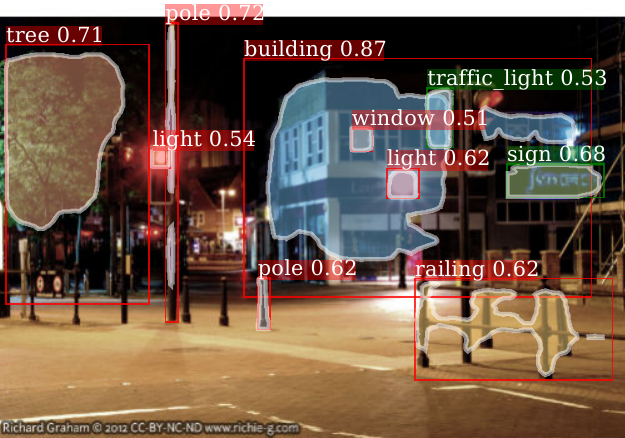} \\
\includegraphics[width=0.24\linewidth]{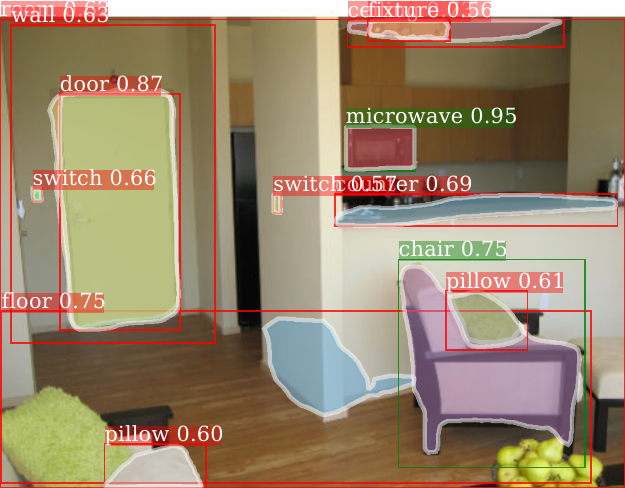}
\includegraphics[width=0.24\linewidth]{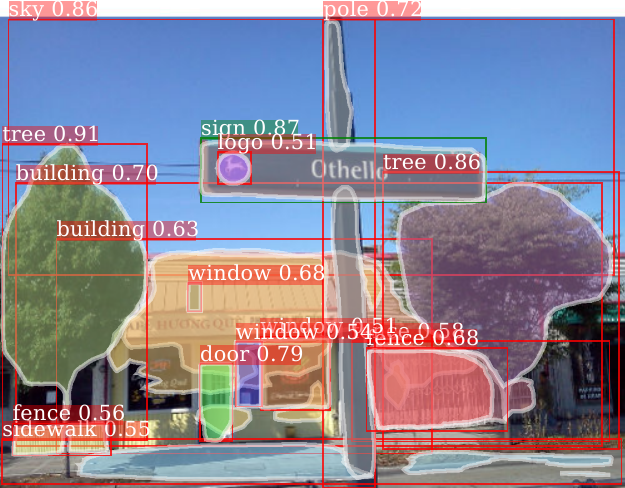}
\includegraphics[width=0.24\linewidth]{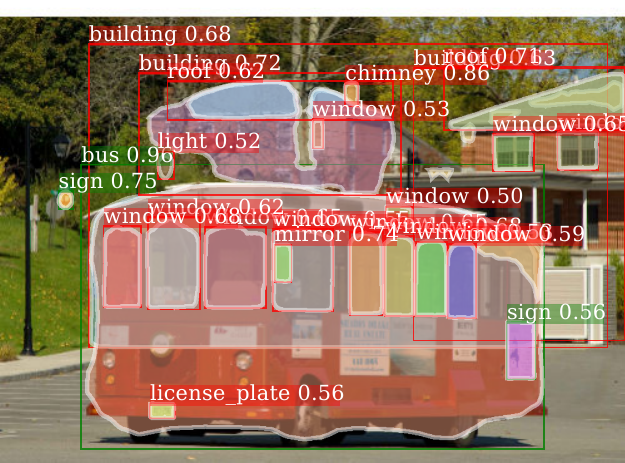}
\includegraphics[width=0.24\linewidth]{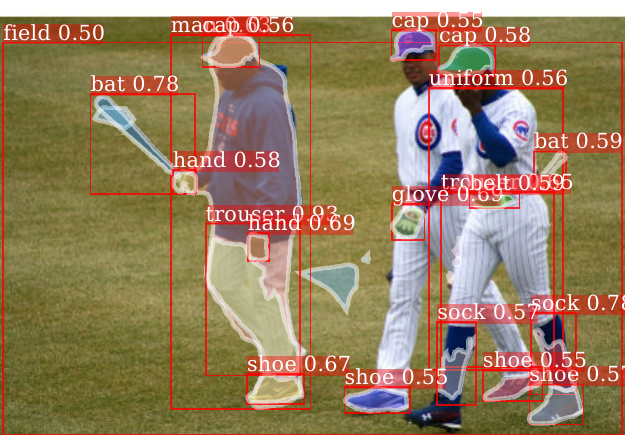} \\
\end{center}
\vspace{-1em}
\caption{\textbf{Example mask predictions from our \methodname on 3000 classes in Visual Genome.} The \textcolor{green}{green} boxes are the 80 classes that overlap with COCO (set $A$ with mask training data) while the \textcolor{red}{red} boxes are the remaining 2920 classes not in COCO (set $B$ without mask training data). It can be seen that our model generates reasonable mask predictions on many classes in set $B$. See \S\ref{sec:exp_vg} for details.}
\label{fig:vis_vg}
\vspace{-1.5em}
\end{figure*}

\subsecvspace
\subsection{Results and Comparison of Our Full Method}
\label{sec:final_comp}
\subsecvspace

Table \ref{tab:result_coco_e2e} compares our full \methodname method (\ie, Mask R-CNN with `transfer+MLP' and $\ftrans$ implemented as `cls+box, 2-layer, LeakyReLU') and the class-agnostic baseline using end-to-end training. In addition, we also compare with the following baseline approaches: a) unsupervised mask prediction using GrabCut \cite{rother2004grabcut} foreground segmentation over the Faster R-CNN detected object boxes (\textbf{Faster R-CNN tested w/ GrabCut}) and b) weakly supervised \textit{instance segmentation} similar to \cite{khoreva2017simple}, which trains an instance segmentation method (here we use Mask R-CNN) on the GrabCut segmentation of the ground-truth boxes (\textbf{Mask R-CNN trained w/ GrabCut}).

\methodname outperforms these approaches by a large margin (over 20\% relative increase in mask AP). We also experiment with ResNet-101-FPN as the backbone network in the bottom half of Table \ref{tab:result_coco_e2e}. The trends observed with ResNet-50-FPN generalize to ResNet-101-FPN, demonstrating independence of the particular backbone used thus far. Figure \ref{fig:vis_coco} shows example mask predictions from the class-agnostic baseline and our approach.

\secvspace
\section{Large-Scale Instance Segmentation}
\label{sec:exp_vg}
\secvspace

Thus far, we have experimented with a simulated version of our true objective: training large-scale instance segmentation models with broad visual comprehension. We believe this goal represents an exciting new direction for visual recognition research and that to accomplish it some form of learning from partial supervision may be required.

To take a step towards this goal, we train a large-scale \methodname model following the partially supervised task, using bounding boxes from the Visual Genome (VG) dataset \cite{krishna2017visual} and instance masks from the COCO dataset \cite{Lin2014}. The VG dataset contains 108077 images, and over 7000 category synsets annotated with object bounding boxes (but not masks). To train our model, we select the 3000 most frequent synsets as our set of classes $A \cup B$ for instance segmentation, which covers all the 80 classes in COCO. Since the VG dataset images have a large overlap with COCO, when training on VG we take all the images that are not in COCO \texttt{val2017} split as our training set, and validate our model on the rest of VG images. We treat all the 80 VG classes that overlap with COCO as our set $A$ with mask data, and the remaining 2920 classes in VG as our set $B$ with only bounding boxes.

\myparagraph{Training.} We train our large-scale \methodname model using the stage-wise training strategy. Specifically, we train a Faster R-CNN model to detect the 3000 classes in VG using ResNet-101-FPN as our backbone network following the hyper-parameters in \S\ref{sec:exp_coco}. Then, in the second stage, we add the mask head using our weight transfer function $\ftrans$ and the class-agnostic MLP mask prediction (\ie, `transfer+MLP'), with the `cls+box, 2-layer, LeakyReLU' implementation of $\ftrans$. The mask head is trained on subset of 80 COCO classes (set $A$) using the mask annotations in the \texttt{train2017} split of the COCO dataset.

\myparagraph{Qualitative results.}
Mask AP is difficult to compute on VG because it contains only box annotations. Therefore we visualize results to understand the performance of our model trained on all the 3000 classes in $A\cup B$ using our weight transfer function. Figure \ref{fig:vis_vg} shows mask prediction examples on validation images, where it can be seen that our model predicts reasonable masks on those VG classes not overlapping with COCO (set $B$, shown in red boxes).

This visualization shows several interesting properties of our large-scale instance segmentation model. First, it has learned to detect abstract concepts, such as shadows and paths. These are often difficult to segment. Second, by simply taking the first 3000 synsets from VG, some of the concepts are more `stuff' like than `thing' like. For example, the model does a reasonable job segmenting isolated trees, but tends to fail at segmentation when the detected `tree' is more like a forest. Finally, the detector does a reasonable job at segmenting whole objects and parts of those objects, such as windows of a trolley car or handles of a refrigerator. Compared to a detector trained on 80 COCO categories, these results illustrate the exciting potential of systems that can recognize and segment thousands of concepts.

\secvspace
\section{Conclusion}
\secvspace

This paper addresses the problem of large-scale instance segmentation by formulating a partially supervised learning paradigm in which only a subset of classes have instance masks during training while the rest have box annotations. We propose a novel transfer learning approach, where a learned \emph{weight transfer function} predicts how each class should be segmented based on parameters learned for detecting bounding boxes. Experimental results on the COCO dataset demonstrate that our method greatly improves the generalization of mask prediction to categories without mask training data. Using our approach, we build a large-scale instance segmentation model over 3000 classes in the Visual Genome dataset. The qualitative results are encouraging and illustrate an exciting new research direction into large-scale instance segmentation. They also reveal that scaling instance segmentation to thousands of categories, without full supervision, is an extremely challenging problem with ample opportunity for improved methods.

{\small
\bibliographystyle{ieee}
\bibliography{everything}

\begin{thebibliography}{10}\itemsep=-1pt

\bibitem{adelson2001seeing}
E.~H. Adelson.
\newblock On seeing stuff: the perception of materials by humans and machines.
\newblock In {\em Human Vision and Electronic Imaging}, 2001.

\bibitem{Bai2017}
M.~Bai and R.~Urtasun.
\newblock Deep watershed transform for instance segmentation.
\newblock In {\em CVPR}, 2017.

\bibitem{bearman2016s}
A.~Bearman, O.~Russakovsky, V.~Ferrari, and L.~Fei-Fei.
\newblock What’s the point: Semantic segmentation with point supervision.
\newblock In {\em ECCV}, 2016.

\bibitem{dai2015boxsup}
J.~Dai, K.~He, and J.~Sun.
\newblock Boxsup: Exploiting bounding boxes to supervise convolutional networks
  for semantic segmentation.
\newblock In {\em ICCV}, 2015.

\bibitem{Dai2015}
J.~Dai, K.~He, and J.~Sun.
\newblock Convolutional feature masking for joint object and stuff
  segmentation.
\newblock In {\em CVPR}, 2015.

\bibitem{Dai2016}
J.~Dai, K.~He, and J.~Sun.
\newblock Instance-aware semantic segmentation via multi-task network cascades.
\newblock In {\em CVPR}, 2016.

\bibitem{dumoulin2016learned}
V.~Dumoulin, J.~Shlens, M.~Kudlur, A.~Behboodi, F.~Lemic, A.~Wolisz,
  M.~Molinaro, C.~Hirche, M.~Hayashi, E.~Bagan, et~al.
\newblock A learned representation for artistic style.
\newblock {\em arXiv preprint arXiv:1610.07629}, 2016.

\bibitem{elhoseiny2013write}
M.~Elhoseiny, B.~Saleh, and A.~Elgammal.
\newblock Write a classifier: Zero-shot learning using purely textual
  descriptions.
\newblock In {\em ICCV}, 2013.

\bibitem{Everingham2010}
M.~Everingham, L.~Van~Gool, C.~K. Williams, J.~Winn, and A.~Zisserman.
\newblock {The PASCAL Visual Object Classes (VOC) Challenge}.
\newblock {\em IJCV}, 2010.

\bibitem{Girshick2014}
R.~Girshick, J.~Donahue, T.~Darrell, and J.~Malik.
\newblock Rich feature hierarchies for accurate object detection and semantic
  segmentation.
\newblock In {\em CVPR}, 2014.

\bibitem{ha2016hypernetworks}
D.~Ha, A.~Dai, and Q.~V. Le.
\newblock {HyperNetworks}.
\newblock {\em arXiv preprint arXiv:1609.09106}, 2016.

\bibitem{Hariharan2014}
B.~Hariharan, P.~Arbel{\'a}ez, R.~Girshick, and J.~Malik.
\newblock Simultaneous detection and segmentation.
\newblock In {\em ECCV}. 2014.

\bibitem{Hariharan2015}
B.~Hariharan, P.~Arbel{\'a}ez, R.~Girshick, and J.~Malik.
\newblock Hypercolumns for object segmentation and fine-grained localization.
\newblock In {\em CVPR}, 2015.

\bibitem{Hayder2017}
Z.~Hayder, X.~He, and M.~Salzmann.
\newblock Boundary-aware instance segmentation.
\newblock In {\em CVPR}, 2017.

\bibitem{He2017}
K.~He, G.~Gkioxari, P.~Doll{\'a}r, and R.~Girshick.
\newblock Mask {R-CNN}.
\newblock In {\em ICCV}, 2017.

\bibitem{He2016}
K.~He, X.~Zhang, S.~Ren, and J.~Sun.
\newblock Deep residual learning for image recognition.
\newblock In {\em CVPR}, 2016.

\bibitem{hoffman2014lsda}
J.~Hoffman, S.~Guadarrama, E.~S. Tzeng, R.~Hu, J.~Donahue, R.~Girshick,
  T.~Darrell, and K.~Saenko.
\newblock {LSDA}: Large scale detection through adaptation.
\newblock In {\em NIPS}, 2014.

\bibitem{khoreva2017simple}
A.~Khoreva, R.~Benenson, J.~Hosang, M.~Hein, and B.~Schiele.
\newblock Simple does it: Weakly supervised instance and semantic segmentation.
\newblock In {\em CVPR}, 2017.

\bibitem{Kirillov2017}
A.~Kirillov, E.~Levinkov, B.~Andres, B.~Savchynskyy, and C.~Rother.
\newblock {InstanceCut}: from edges to instances with multicut.
\newblock In {\em CVPR}, 2017.

\bibitem{krishna2017visual}
R.~Krishna, Y.~Zhu, O.~Groth, J.~Johnson, K.~Hata, J.~Kravitz, S.~Chen,
  Y.~Kalantidis, L.-J. Li, D.~A. Shamma, M.~Bernstein, and L.~Fei-Fei.
\newblock Visual genome: Connecting language and vision using crowdsourced
  dense image annotations.
\newblock {\em {IJCV}}, 2017.

\bibitem{Li2017}
Y.~Li, H.~Qi, J.~Dai, X.~Ji, and Y.~Wei.
\newblock Fully convolutional instance-aware semantic segmentation.
\newblock In {\em CVPR}, 2017.

\bibitem{Lin2017}
T.-Y. Lin, P.~Doll{\'a}r, R.~Girshick, K.~He, B.~Hariharan, and S.~Belongie.
\newblock Feature pyramid networks for object detection.
\newblock In {\em CVPR}, 2017.

\bibitem{Lin2014}
T.-Y. Lin, M.~Maire, S.~Belongie, J.~Hays, P.~Perona, D.~Ramanan,
  P.~Doll{\'a}r, and C.~L. Zitnick.
\newblock {Microsoft COCO}: Common objects in context.
\newblock In {\em ECCV}, 2014.

\bibitem{Long2015}
J.~Long, E.~Shelhamer, and T.~Darrell.
\newblock Fully convolutional networks for semantic segmentation.
\newblock In {\em CVPR}, 2015.

\bibitem{maas2013rectifier}
A.~L. Maas, A.~Y. Hannun, and A.~Y. Ng.
\newblock Rectifier nonlinearities improve neural network acoustic models.
\newblock In {\em ICML}, volume~30, 2013.

\bibitem{mikolov2013efficient}
T.~Mikolov, K.~Chen, G.~Corrado, and J.~Dean.
\newblock Efficient estimation of word representations in vector space.
\newblock {\em arXiv preprint arXiv:1301.3781}, 2013.

\bibitem{misra2017red}
I.~Misra, A.~Gupta, and M.~Hebert.
\newblock From red wine to red tomato: Composition with context.
\newblock In {\em CVPR}, 2017.

\bibitem{Pan2010}
S.~J. Pan and Q.~Yang.
\newblock A survey on transfer learning.
\newblock {\em IEEE Trans. on Knowl. and Data Eng.}, 2010.

\bibitem{papandreou2015weakly}
G.~Papandreou, L.-C. Chen, K.~Murphy, and A.~L. Yuille.
\newblock Weakly-and semi-supervised learning of a {DCNN} for semantic image
  segmentation.
\newblock In {\em ICCV}, 2015.

\bibitem{pathak2015constrained}
D.~Pathak, P.~Krahenbuhl, and T.~Darrell.
\newblock Constrained convolutional neural networks for weakly supervised
  segmentation.
\newblock In {\em ICCV}, 2015.

\bibitem{pennington2014glove}
J.~Pennington, R.~Socher, and C.~Manning.
\newblock {GloVe}: Global vectors for word representation.
\newblock In {\em {EMNLP}}, 2014.

\bibitem{Pinheiro2015}
P.~O. Pinheiro, R.~Collobert, and P.~Dollar.
\newblock Learning to segment object candidates.
\newblock In {\em NIPS}, 2015.

\bibitem{Pinheiro2016}
P.~O. Pinheiro, T.-Y. Lin, R.~Collobert, and P.~Doll{\'a}r.
\newblock Learning to refine object segments.
\newblock In {\em ECCV}, 2016.

\bibitem{Ren2015a}
S.~Ren, K.~He, R.~Girshick, and J.~Sun.
\newblock {Faster R-CNN}: Towards real-time object detection with region
  proposal networks.
\newblock In {\em NIPS}, 2015.

\bibitem{rother2004grabcut}
C.~Rother, V.~Kolmogorov, and A.~Blake.
\newblock {GrabCut}: Interactive foreground extraction using iterated graph
  cuts.
\newblock In {\em ACM ToG}, 2004.

\bibitem{Russakovsky2015}
O.~Russakovsky, J.~Deng, H.~Su, J.~Krause, S.~Satheesh, S.~Ma, Z.~Huang,
  A.~Karpathy, A.~Khosla, M.~Bernstein, A.~C. Berg, and L.~Fei-Fei.
\newblock {ImageNet Large Scale Visual Recognition Challenge}.
\newblock {\em IJCV}, 2015.

\bibitem{tsai2017learning}
Y.-H.~H. Tsai, L.-K. Huang, and R.~Salakhutdinov.
\newblock Learning robust visual-semantic embeddings.
\newblock {\em arXiv preprint arXiv:1703.05908}, 2017.

\bibitem{wang2016learning}
Y.-X. Wang and M.~Hebert.
\newblock Learning to learn: Model regression networks for easy small sample
  learning.
\newblock In {\em ECCV}, 2016.

\bibitem{xian2016latent}
Y.~Xian, Z.~Akata, G.~Sharma, Q.~Nguyen, M.~Hein, and B.~Schiele.
\newblock Latent embeddings for zero-shot classification.
\newblock In {\em CVPR}, 2016.

\end{thebibliography}
}

\end{document}